\documentclass[10pt,twocolumn,letterpaper]{article}

\usepackage{iccv}
\usepackage{times}
\usepackage{epsfig}
\usepackage{graphicx}
\usepackage{amsmath}
\usepackage{amssymb}
\usepackage{amsthm}
\usepackage[final]{pdfpages} 

\usepackage{booktabs}
\usepackage[style=base]{subcaption}

\usepackage{pgfplots}
\pgfplotsset{compat=1.9}
\usepgfplotslibrary{external}

\usepackage[pagebackref=true,breaklinks=true,colorlinks,bookmarks=false]{hyperref}

\iccvfinalcopy 

\def\iccvPaperID{} 

\graphicspath{{./figs/}}

\newif\ifpgf
\pgffalse   

\newtheorem{defi}{Definition}
\newtheorem{prop}{Proposition}
\newtheorem{rem}{Remark}

\def\R{\mathbb{R}}

\newcommand{\X}{\mathcal X}

\newcommand{\F}{\mathcal F}
\renewcommand{\H}{\mathcal H}
\renewcommand{\O}{\mathcal O}
\newcommand{\G}{\mathcal G}
\newcommand{\T}{\mathcal T}
\newcommand{\C}{\mathcal C}

\newcommand{\nor}[1]{\left\lVert {#1} \right\rVert}

\newcommand{\ind}{1{\hskip -2.5 pt}\hbox{I}}

\renewcommand\iff{\Leftrightarrow} 
\newcommand{\capmar}{\vspace{-0.3cm}} 

\begin{document}
\null
\includepdf{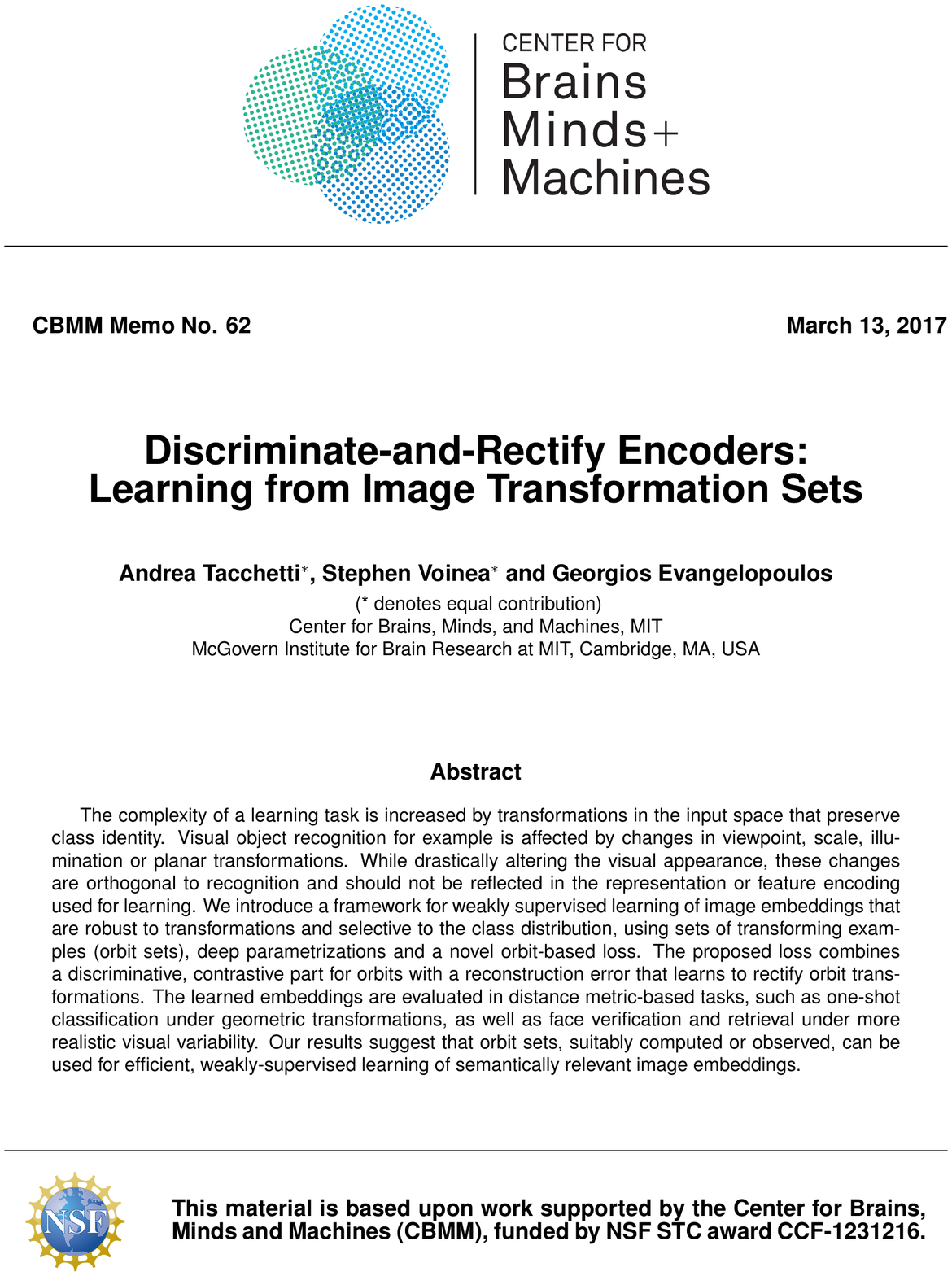}
\setcounter{page}{1}

\title{Discriminate-and-Rectify Encoders: Learning from Image Transformation Sets}

\author{
    Andrea Tacchetti*, Stephen Voinea* and Georgios Evangelopoulos\\
    The Center for Brains, Minds and Machines, MIT | McGovern Institute for Brain Research at MIT,\\
    {\tt\small \{atacchet, voinea, gevang\}@mit.edu}\\
    {\small *denotes equal contribution}
}

\maketitle

\begin{abstract}
The complexity of a learning task is increased by transformations in the input space that preserve class identity. Visual object recognition for example
is affected by changes in viewpoint, scale, illumination or planar transformations. While drastically altering the visual appearance, these changes are orthogonal to recognition and should not be reflected in the representation or feature encoding used for learning. We introduce a framework for weakly supervised learning of image embeddings that are robust to transformations and selective to the class distribution, using sets of transforming examples (orbit sets), deep parametrizations and a novel orbit-based loss. The proposed loss combines a discriminative, contrastive part for orbits with a reconstruction error that learns to rectify orbit transformations. The learned embeddings are evaluated in distance metric-based tasks, such as one-shot classification under geometric transformations, as well as face verification and retrieval under more realistic visual variability. Our results suggest that orbit sets, suitably computed or observed, can be used for efficient, weakly-supervised learning of semantically relevant image embeddings.
\end{abstract}

\section{Introduction}

The distribution of examples for a learning problem, such as visual object recognition, will exhibit variability across and within semantic categories. The former is due to the category-specific statistics; the latter is due to the variety of instances that share the same semantics 
and by transformations that preserve the identity, such as geometric or photometric changes. Such transformations will alter the properties of the visual scene but will not change the semantic category of an object. Recognition across novel views (position, size, pose), clutter and occlusions \cite{DiCarlo2012,Poggio2016,Leibo2015}, and generalization to new examples from a category,
are hallmarks of human and primate perception. Invariance to transformations has been consistently explored as the computational objective of representations for computer vision and machine learning \cite{Hadsell2006,Hinton2011,Bruna2013,Lenc2015,Soatto2016,Cohen2016}.

\begin{figure}[!t]
    \centerline{
        \includegraphics[width=\columnwidth]{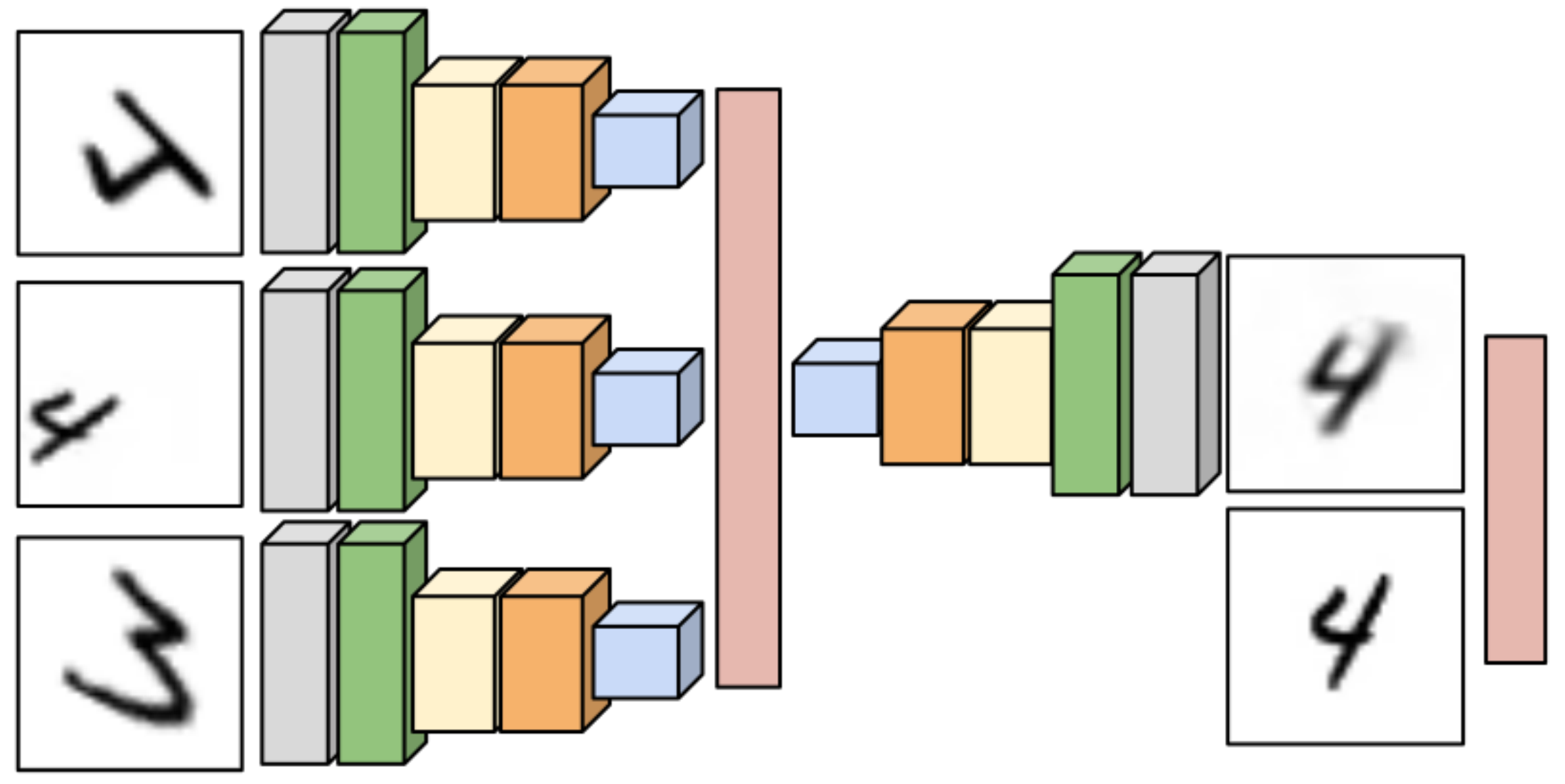}    }
    \capmar
    \caption{\small Discriminate-and-Rectify Encoder Network: Convolutional layers are followed by max pooling (cascade of two shown here). Orbit-Triplet (OT) loss is using only the encoding part for a triplet. Orbit-Encode (OE) reconstructs a canonical orbit element (lower right) from the deconvolutional decoder output. Tied weights are denoted by same colors, loss layers by red.}
    \label{fig:architecture}
\end{figure}

Representations that facilitate generalization in downstream supervised tasks can be learned in unsupervised or semi-supervised settings \cite{Anselmi2015b}, where the distribution of observations is used for obtaining non-linear similarity metrics, reducing the dimensionality by disregarding nuisance directions or deriving interpretable, generative models. Unsupervised learning has been used, for example, 
for pre-training of neural networks with the goal of improving the convergence rates of end-to-end learning algorithms. 

A question of theoretical and practical interest is under which conditions can representations learned without explicit supervision \cite{Bengio2013} match the performance of supervised learning methods that implicitly account for the representation of the data \cite{LeCun2015}. Learning from unlabeled (or implicitly labeled) data can be an alternative to using labeled examples for training multi-parameter deep neural networks, and building large-scale, generic and transferable learning systems economically. In addition, biological and cognitive learning paradigms, particularly in perceptual domains such as vision or speech, predict learning and generalization from a small number of labeled examples and an abundance of implicitly labeled observations or weak supervision.

A natural source of weak supervision is the formation of equivalence relations and classes in the input space, that are not necessarily related to the learning task. Such relations can be, for example, temporal, categorical or generative and partition the space in sets 
which we will loosely refer to as obits in this paper. Example of orbits are the set of images of an object under rotations \cite{Anselmi2015b,Hadsell2006} or the frames of a video of a moving object \cite{Mobahi2009,Zou2012,Memisevic2013,Wang2015}. This partition by orbit sets, in terms of granularity, lies in-between single-example and task-specific, semantic class partitions.  

In this paper, we propose using orbit sets, defined by generic transformations, as weak supervision for learning representations in an invariant metric space. Two points are equivalent if one can be related to the other through a transformation, and the set of all equivalent points forms an orbit. Orbits are either explicitly generated (data augmentation) or implicitly specified (temporal continuity, data acquisition or association). As opposed to inference using an explicit transformation model \cite{Jaderberg2015} or factoring out the nuisance using explicit pooling \cite{Anselmi2015b}, we learn deep, parametric embeddings using a novel loss function that incorporates the orbit equivalence relations.

The proposed \emph{orbit metric loss} generalizes the triplet loss and denoising autoencoders to orbit sets, that promote approximate invariance and reconstruction. We separately study the two motivating special cases: the \emph{orbit triplet loss}, a discriminative term which implicitly promotes invariance and selectivity with respect to examples drawn from the same or different orbits respectively, and the \emph{orbit encoder loss}, a generative term, which learns to rectify or \emph{de-transform} by mapping orbit points to a single, canonical element. The learned embeddings are compared, under the same parametrizations, to those from the supervised triplet loss \cite{Schroff2015}, the surrogate class loss \cite{Dosovitskiy2015}
and, when a full model of the orbit-generating process is available (e.g. affine transformations), spatial transformer networks \cite{Jaderberg2015}. 

The learned embeddings define robust metrics that are semantically relevant for distance-based and low supervised-sample regime tasks, such as ranking, retrieval, matching, clustering, graph-construction and relational learning.
We provide
quantitative comparisons on
face verification and retrieval (on Multi-PIE dataset) and one-shot learning for classification (on MNIST with affine transformations). Our results show that partitioning the input space according to suitable orbit sets is a powerful weak supervision cue, which the proposed encoding loss can exploit effectively
to learn semantically relevant embeddings.

\section{Related work}

Invariance to transformations that are orthogonal to the learning task has been the subject of extensive theoretical and empirical investigation in artificial and biological perception and recognition. A number of studies focused on theoretical insights on the trade-off between invariance and selectivity through sufficient statistics \cite{Soatto2016}, the optimality of explicit parametrizations with convolutions/pooling and memory-based learning \cite{Anselmi2015b,Anselmi2016}, as well as constructing invariants for compact groups and maps that are robust, rather than invariant, to diffeomorphisms \cite{Mallat2012}.
These inspired feature extraction architectures for object recognition \cite{Dong2015}, texture classification \cite{Bruna2013}, face verification \cite{Liao2013}, action recognition \cite{Tacchetti2016}, and speech recognition \cite{Voinea2014,Zhang2015}. 

In this paper, we rely on the theoretical framework in \cite{Anselmi2015b,Anselmi2016}. While relaxing some of the assumptions, e.g. compact groups, exact invariance, we make use of generic orbit sets, that can come from implicit supervision \cite{Anselmi2015b}, or include non-group transformations, partial orbits and noisy samples. We propose a loss function that is a proxy for a margin-based invariance and selectivity in the representation, which can be used for end-to-end trainable encoders and does not rely on a particular parametrization, nor does it require access to a model of the orbit generation process.

Side information as a form of weak supervision has been employed for various distance metric learning algorithms \cite{Xing2003};
for example, variants of the triplet loss function which rely on knowing which of two pairs corresponds to similar samples \cite{Weinberger2009,Song2016}. Supervised versions of the triplet loss have have been used for discriminatively-trained metric learning, through convolutional network parametrization, aiming to minimize the true objective of the task (e.g., face verification) \cite{Chopra2005, Schroff2015}. The triplet loss was also used for nonlinear dimensionality reduction and learning transformation-invariant embeddings \cite{Hadsell2006}. Similar to the neighborhood graphs in \cite{Hadsell2006}, the orbits in our work can be obtained by side information or temporal proximity and are assumed known only for training. 

Representation learning through surrogate classes populated by data augmentation transformations was explored in the exemplar CNN framework \cite{Dosovitskiy2015}. Autoencoder networks \cite{Hinton2006,Baldi2012,Bengio2013} have been typically used for dimensionality reduction (bottleneck features) and unsupervised pre-training of deep networks. Different reconstruction requirements or regularization terms have lead to useful encodings by learning to perform denoising \cite{Vincent2008}, sparse coding, contractive approximations for robustness \cite{Rifai2011}, or respect temporal continuity (feature slowness) \cite{Zou2012,Mobahi2009}. Convolutional autoencoders enforce local spatial robustness through max pooling \cite{Zhao2016,Masci2011}. Our method uses a combination of triplet loss and an explicit rectification accuracy loss through an encoder-decoder network.

The idea of using a representation of the transformations for robust embeddings has been explored through explicit estimation of the parameters of an exact model of the generative process in spatial transformer networks \cite{Jaderberg2015}, estimation of a latent representation in transforming autoencoders \cite{Hinton2011}, or a distributed representation \cite{Zhao2016}. Similar to \cite{Sajjadi2016}, the proposed orbit loss functions 
can be used as a regularizer of the discriminative loss of a deep network for representations that are both robust and discriminatively trained.  

\section{Background}

We begin by reviewing relevant background concepts in order to provide context and formulate clearly the proposed losses and weakly-supervised learning methods.

\noindent \textbf{Learning and representations}
The feature map or data representation $\Phi\!:\!\X \rightarrow \F$ for a learning problem on input space $\X$, can be explicitly selected or learned using principles such as 
\begin{itemize}
    \item distance preservation $|| \Phi(x) - \Phi(x') ||_{\F} \approx \nor{x - x'}$ or contraction $|| \Phi(x) - \Phi(x') ||_{\F} \leq l \nor{x - x'}, l \in[0,1)$, 
    \item reconstruction $ \nor{x - \tilde{\Phi} \circ \Phi(x)} \leq \epsilon, \tilde{\Phi}\!:\!\F \rightarrow \X$, 
    \item invariance and selectivity $x' \sim x \iff \Phi(x) = \Phi(x')$,.
\end{itemize}
where $\sim$ denotes equivalence and $\circ$ function composition, i.e. $(\tilde{\Phi} \circ \Phi)(x) = \tilde{\Phi}(\Phi(x))$. 

{\rem The feature map $\Phi$ is selected from some hypothesis space of functions $ \H \subseteq \{ \Phi\,|\,\Phi\!:\!\X \rightarrow \F\} $. In practice, some parametrization of the elements of $\H$ that renders the resulting representation learning problem tractable is necessary.}
{\rem For kernel machines (or shallow networks), $\Phi$ is preselected, or implicitly induced by a kernel function $K: \X \times \X \rightarrow \R$ , $K(x,x') = \langle \Phi(x), \Phi(x')\rangle, \forall x, x' \in \X$. For deep networks, $\Phi$ is parametrized, typically through linear projections and non-linearities, and jointly learned with the predictor function. 
Moreover, it involves multiple maps $\Phi_l\!:\!\F_{l-1} \rightarrow \F_l$ in the form of compositions of multiple representation layers $\Phi = \bar{\Phi}_L \circ \cdots \circ \bar{\Phi}_1$.}\\

\noindent \textbf{Metric learning} In the general case,
the global metric learning problem \cite{Kulis2013} is learning a distance function $D(x, x^\prime)$ between two points $x, x^\prime \in \X$ as the distance $D_\F\left(\Phi(x), \Phi(x^\prime)\right)$ in a new space $\F$:
\begin{equation}
    D(x, x^\prime) = (\Phi(x) - \Phi(x^\prime))^T(\Phi(x) - \Phi(x^\prime)).
    \label{eq:metric}
\end{equation} 
The representation $\Phi: \X \rightarrow \F$ can express linear, kernelized or nonlinear mappings and is obtained by the solution of a regularized, constrained minimization problem, using some form of side information, such as the similarity between pairs \cite{Xing2003} or triplets of points \cite{Song2016}.\\

\noindent \textbf{Triplet loss}
The large margin nearest neighbor loss \cite{Weinberger2009} was developed for supervised learning of a distance metric by pulling together and pushing apart same- and different-class neighbors, respectively. The closely related contrastive loss \cite{Chopra2005} uses pairs of observations $(x_i, x_j)$ and their label $(y_i, y_j)$ agreement to decrease or increase their distance $D(x_i, x_j) = \nor{\Phi(x_i) - \Phi(x_j)}^2_{\F}$ by learning $\Phi$ through:         
\begin{equation} 
    \min_{ \Phi \in \H} \sum_{i,j\neq i}^n \ind_{y_i = y_j}D(x_i, x_j) +  (1-\ind_{y_i=y_j})\left|\alpha  - D(x_i, x_j)\right|_{+}, 
\end{equation}
where $n$ is the size of the training set $\{(x_i, y_i)\}_{1}^{n}$, $\alpha \in \R_+$ a distance margin for the non-matching pairs and $\left|\alpha\right|_{+}=\max\{0,\alpha\}$ the hinge loss function. The triplet loss is based on defining point triplets using their label agreement $\T = \{(x_i, x_p, x_q)\, |\, y_i=y_p, y_i \neq y_q\}$ \cite{Schroff2015} and aims to enforce
\begin{equation}
    D(x_i, x_p) \leq D(x_i, x_q) - \alpha
    \label{eq:triplet_motiv}
\end{equation}
by minimizing the mismatch part of the large margin loss:
\begin{equation} 
    \min_{\Phi \in \H} \sum_{i=1}^{|\T|} \left|D(x_i, x_p)  +  \alpha  - D(x_i, x_q) \right|_+. 
    \label{eq:triplet_loss}
\end{equation}

\noindent \textbf{Autoencoders} An autoencoder is composed of an encoding map $\Phi\!:\!\R^d \rightarrow \R^k$ and a decoding map $\tilde{\Phi}\!:\!\R^k \rightarrow \R^d$, where we assume $\X$ and $\F$ be Euclidean spaces and $\H$ and $\tilde{\H}$ to be the appropriate hypothesis spaces, 
learned by minimizing a reconstruction loss
\begin{equation} 
    \min_{\Phi \in \H, \tilde{\Phi} \in \tilde{\H}} \sum_{i=1}^n  L(x_i, \tilde{\Phi}\circ\Phi(x_i))
\end{equation}
where
$L:\R^d \times \R^d \rightarrow \R_+$ is typically the square or cross-entropy loss. The encoding is parametrized by a linear map using $k$ projection units or filters $W = (w_1, \ldots, w_k) \in \R^{d \times k}$, an offset $b \in \R^d$, and a nonlinear function $\sigma:\R \rightarrow \R$ applied element-wise  
\begin{equation}
    \Phi(x) = \sigma \circ (Wx + b). \label{eq:enc_basic}
\end{equation}
The decoding map is typically of a similar form 
\begin{equation} 
    \tilde{\Phi} \circ \Phi(x) = \sigma \circ (\widetilde{W}\Phi(x) + \tilde{b})
    \label{eq:dec_basic}
\end{equation} 
usually constrained having tied-weights, $\widetilde{W} = W^T$, for a reduced number of parameters. Both maps can have multiple layers and learned with additional priors through regularization on $W$ or the activations of the hidden layers \cite{Bengio2013,Rifai2011}, reconstructing perturbations of $x$ \cite{Vincent2008} or convolutional structure on $W$ \cite{Zeiler2010,Masci2011,Zhao2016} and pooling.\\  

\begin{figure}[t!]
    \captionsetup[subfigure]{labelformat=empty}
    \begin{subfigure}{.125\linewidth}
        \centering
        \includegraphics[width=\linewidth]{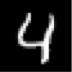}
    \end{subfigure}%
    \begin{subfigure}{.125\linewidth}
        \centering
        \includegraphics[width=\linewidth]{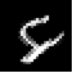}
    \end{subfigure}%
    \begin{subfigure}{.125\linewidth}
        \centering
        \includegraphics[width=\linewidth]{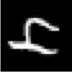}
    \end{subfigure}%
    \begin{subfigure}{.125\linewidth}
        \centering
        \includegraphics[width=\linewidth]{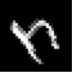}
    \end{subfigure}%
    \begin{subfigure}{.125\linewidth}
        \centering
        \includegraphics[width=\linewidth]{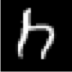}
    \end{subfigure}%
    \begin{subfigure}{.125\linewidth}
        \centering
        \includegraphics[width=\linewidth]{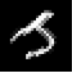}
    \end{subfigure}%
    \begin{subfigure}{.125\linewidth}
        \centering
        \includegraphics[width=\linewidth]{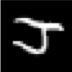}
    \end{subfigure}%
    \begin{subfigure}{.125\linewidth}
        \centering
        \includegraphics[width=\linewidth]{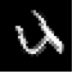}
    \end{subfigure}%

    \begin{subfigure}{.125\linewidth}
        \centering
        \includegraphics[width=\linewidth]{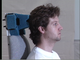}
    \end{subfigure}%
    \begin{subfigure}{.125\linewidth}
        \centering
        \includegraphics[width=\linewidth]{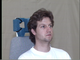}
    \end{subfigure}%
    \begin{subfigure}{.125\linewidth}
        \centering
        \includegraphics[width=\linewidth]{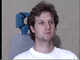}
    \end{subfigure}%
    \begin{subfigure}{.125\linewidth}
        \centering
        \includegraphics[width=\linewidth]{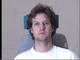}
    \end{subfigure}%
    \begin{subfigure}{.125\linewidth}
        \centering
        \includegraphics[width=\linewidth]{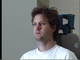}
    \end{subfigure}%
    \begin{subfigure}{.125\linewidth}
        \centering
        \includegraphics[width=\linewidth]{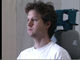}
    \end{subfigure}%
    \begin{subfigure}{.125\linewidth}
        \centering
        \includegraphics[width=\linewidth]{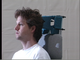}
    \end{subfigure}%
    \begin{subfigure}{.125\linewidth}
        \centering
        \includegraphics[width=\linewidth]{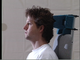}
    \end{subfigure}%
    \capmar
    \caption{\small Examples of transformation orbits obtained through image rotations/data augmentation (row 1: MNIST, digits) and continuity (row 2: MulitPIE, faces).}
    \label{fig:orbex}
\end{figure}


\noindent \textbf{Transformations and orbit sets}
Consider a family of transformations as a set of maps $\G \subset \{g\,|\, g: \X \rightarrow \X\}$. We will denote by $gx$ the action of the transformation represented by $g$ on $x$, which generates point $x'$, i.e. $x' = gx = g(x)$. The transformations can be parametrized by $\theta_j \in \Theta$, such that $\G = \{g_j = g_{\theta_j}\,|\, \theta_j \in \Theta\}$. The set can have algebraic structure, e.g. form a group \cite{Anselmi2016, Mallat2012, Cohen2016} (Fig.~\ref{fig:orbex}, row 1). 

{\defi[Group orbits \cite{Anselmi2016}] An orbit associated to an element $x \in \X$ is the set of points that can be reached under the transformations $\G$, i.e., $\O_x = \{gx \in \X | g\in \G \} \subset \X$.} 
\noindent Given a group structure on $\G$, the transformations partition the input space $\X$ into orbits by defining equivalence relations: $x \sim x' \iff \exists g \in \G: x' = gx, \forall x, x' \in \X $. As a result, each $x \in \X$ belongs to one and only one orbit $\O_{x} = \O_{gx}, \forall g \in \G$ and the input space is $\X = \cup_{x \in \X, g \in \G} \O_{gx} = \cup_{x \in \X} \O_{x}$. Using the fact that orbit are sets defined by equivalence relations in $\X$, we can extend the definition to relations or set memberships provided by categorical labels.
{\defi[Generic orbits] An orbit associated to an element $x \in \X$ is the subset of $\X$ that includes $x$ along with an equivalence relation, i.e. the equivalence class $\O_x = \{x^\prime \in \X | x \sim x^\prime\} \subset \X$. The equivalence relation is given by a function $c: \X \rightarrow \C$ such that $ x \sim x^\prime \iff c(x) = c(x^\prime)$.} 

\noindent Examples of such maps are the labels of a supervised learning task, the indexes of vector quantization codewords or, for the case of sequential data such as videos, the sequence membership, with $\C$ the set of classes, codewords or sequences respectively (Fig.~\ref{fig:orbex}, row 2).

\noindent \textbf{Surrogate classes and exemplar loss} The Exemplar loss, introduced as a way to combine data augmentation and weak supervision for training convolutional networks \cite{Dosovitskiy2015}, uses a surrogate class for each point in an unlabeled training set $\X_n = \{x_i\}_{i=1}^{n}$. The surrogate class instances are generated by random transformations, sampled from $\G$, of the class prototypes $\{g_jx_i\}_{j=1}^{k_i \leq |\G|}$. An embedding $\Phi$ is learned by minimizing a discriminative loss with respect to the surrogate classes: 
\begin{equation}  
     \min_{\Phi \in \H, f\in \H_f} \sum_{i=1}^n \sum_{j=1}^{k_i \leq |G|} L(i, f(\Phi(g_jx_i))) 
    \label{eq:exemplar}
\end{equation} 
where $i$ indexes the original, untransformed training set $\X_n$ and serves as the surrogate class label for all points generated from $x_i$; $f$ is a classifier learnt jointly with the embedding. 

\noindent \textbf{Spatial transformer networks (STNs)} When a plausible forward model of the process that generates orbits is known, i.e. when a suitable parametrization of $\G = \{g_j = g_{\theta_j}\,|\, \theta_j \in \Theta\}$ is available, STNs \cite{Jaderberg2015} are trainable modules that learn to undo a transformation in $\G$, by explicitly transforming the input of a feature map. STNs introduce a specific modification to the parametrization of $\Phi$, that for an input $g_{\theta} x_i$, provides an estimate of $\theta$ and applies the inverse transformation $g_{\tilde{\theta}}\circ g_{\theta}x_i \approx x_i$. This module acts as an {\it oracle} that provides a rectified, untransformed version of its input, which is then passed to downstream embedding maps. The resulting embedding is robust, by construction, to transformations in $\G$.

\begin{figure*}
    \centerline{
        \includegraphics[width=0.5\textwidth]{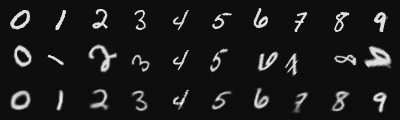}\hspace{-0.087cm}
        \includegraphics[width=0.5\textwidth]{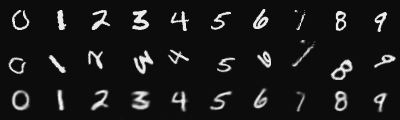}}
    \capmar
    \caption{\small Rectification examples from the orbit decoder output for samples from affine MNIST, showing canonical image $x_c$ (top), randomly transformed $x_i=g_ix_c$ (middle) and rectified output $\tilde{\Phi} \circ \Phi(x_i)$ (bottom).}
    \label{fig:orbenc}
\end{figure*}

\section{Metric learning with orbit loss}

We introduce a novel loss function for learning an embedding for a distance metric using as weak supervision the set memberships on transformation orbits. The loss aims to jointly, adaptively and in a data-driven manner enforce \emph{invariance}, to the transformations captured by the orbit sets, \emph{selectivity} and \emph{low rectification error} on the representation. The loss function is independent of the embedding parametrization, though it implies a siamese (tied weights) and an encoder-decoder network architecture (Fig.~\ref{fig:architecture}). In this paper we will learn deep, convolutional encodings. Orbit sets are obtained either from explicit transformations of the unlabeled input data, e.g. each sample generates an orbit, or from a weak supervision signal involving data continuity, e.g. subsets of the training set correspond to data collected sequentially or under multiple views.  

\subsection{Problem statement} 
Let training set $\X_n = \{x_i\}_{i=1}^n \in \X$ be a set of unlabeled instances. We assume the input to be in $\X=\R^d$, for example having $x_i$ be the vectorized intensity values of an image. We aim to learn a feature map or embedding $\Phi: \X \rightarrow \F$, in space $\F=\R^k$ with $k \leq d$, such that the metric $D: \X \times \X \rightarrow \R_{+}$ given by the distance
\begin{equation}
    D(x, x^\prime) = ||\Phi(x) - \Phi(x^\prime)||^2_{\F}, \label{eq:prob_metric}
\end{equation}
where $||\cdot||_{\F}$ denotes the norm in $\F$, is invariant and selective with respect to the transformations of $x$ captured by an orbit set $\O_x$, equivalently written as: 
\begin{equation} 
    x^\prime \sim x \iff x, x^\prime \in \O_x \iff D(x, x^\prime) \leq \epsilon,
    \label{eq:inv_sel}
\end{equation} 
where we relax the exact invariance condition using an $\epsilon$-approximation to the zero-norm distance. In terms of $\Phi$, Eq.~(\ref{eq:inv_sel}) defines a sufficient and necessary condition for two points being equivalent under the transformation in the orbit set \cite{Anselmi2016}. Note that the requirement for selectivity, i.e., the converse direction, makes $(\R^d, D)$ a proper metric space with an invariant, in this $\epsilon$-approximation sense, metric. 
    
\subsection{Orbit sets}
\label{subsec:orbit_sets}

The definition of the orbit sets $\O_x$ is crucial for the proposed framework and can be based on the data distribution and the learning problem; here we give a few examples of orbit sets. \\
\noindent \textbf{Augmentation} Given a parametrized family of transformations $\G$, one can generate orbit samples for a given $x$ by randomly sampling from the parameter vectors $\{\theta_j \in \Theta\}$ and letting $\O_x = \{g_{\theta_j}x\, | \, \theta_j \in \Theta\}$. Examples include geometric transformations (rotation, translation, scaling), e.g. Fig~\ref{fig:orbex}~(row 1), or typical data augmentation transforms (cropping, contrast, color, blur, illumination etc.) \cite{Dosovitskiy2015}.\\
\noindent \textbf{Acquisition} If the data acquisition process is part of the learning problem, e.g. online/unsupervised learning, or included as meta-data, e.g. multiple samples of an object across time, conditions or views, e.g. Fig~\ref{fig:orbex}~(row 2), then an orbit can be associated to all samples from the same sequence or session \cite{Gross2010}.\\
\noindent \textbf{Temporal continuity} For sequential data such as videos, an orbit can be a continuous segment of the video stream, following plausible assumptions on feature smoothness and continuity of the representation in time \cite{Wang2015, Zou2012}.     
     
\subsection{Orbit metric loss} 
\label{subsec:oj} 

Assume the set of orbits $\{\O_{x_i}\}$ given, either via an a priori partition of the training set $\X_n$ in a number of equivalence classes such that $\X_n = \cup_{x_i}\O_{x_i}$, or by augmentation of each $x_i \in \X_n$ such that $\O_{x_i} = \{g_{\theta_j}x\, | \, \theta_j \in \Theta\}$. Given the orbits, consider a set of triplets 
\begin{equation} 
    \T \subset \{(x_i, x_p, x_q)\, |\, x_i \in \X_n, x_p \in \O_{x_i}, x_q \in \O_{x_q} \neq \O_{x_i} \}
    \label{eq:triplets}
\end{equation} 
such that each $x_i$ is assigned a \emph{positive example} $x_p$ (in-orbit), i.e. $x_i \sim x_p \iff x_i, x_p \in \O_{x_i}$ and a \emph{negative example} $x_q$ (out-of-orbit), i.e. $\O_{x_q} \neq \O_{x_i}$. We further assume that each orbit is equipped with a \emph{canonical example} $x_c \in \O_{x_i}$.  

{\defi[Orbit canonical element] An orbit point that provides a reference coordinate system for the family of transformations that generate the orbit. For $\O_x$ obtained through a generative process applied on $x$, $x_c$ is the output of the identity transformation $g_0 \in \G$, i.e. $x_c = g_0x$.}. For orbits from categorical meta-data, $x_c$ is empirically chosen to be the `regular' view or neutral condition (Fig.~\ref{fig:orbenc}, row 1).

The proposed loss function, reflected in the architecture in Fig.~\ref{fig:architecture}, is composed of two terms; a discriminative term  $L_{{\rm t}}(x_i, x_p, x_q)$, based on the triplet loss, using distances between the encodings $\Phi:\R^d\rightarrow\R^k$ on the feature space $\R^k$; a reconstruction error $L_{{\rm e}}(x_i, x_c)$ between a decoder $\tilde{\Phi}:\R^k\rightarrow\R^d$ output and the canonical, as a distance on the input space $\R^d$:
\begin{equation}
    \begin{gathered} 
        \min_{\Phi, \tilde{\Phi}} \sum_{x_i \in \X_n} \frac{\lambda_1}{d} L_{{\rm t}}(t_i) + \frac{\lambda_2}{k} L_{{\rm e}}(x_i, x_c), \quad t_i = (x_i, x_p, x_q) \\
    L_{{\rm t}}(t_i) =  \left|\nor{\Phi(x_i) - \Phi(x_p)}^2_{\R^k} \!+\! \alpha \!-\! \nor{\Phi(x_i) -     \Phi(x_q)}^2_{\R^k} \right|_+ \\
L_{{\rm e}}(x_i, x_c) = \nor{x_c - \tilde{\Phi}\circ\Phi(x_i)}^2_{\R^d}.
    \end{gathered}
    \label{eq:orbit_loss}
\end{equation}
The constants $\lambda_1$, $\lambda_2$ and $\alpha$ control the relative contribution of each term and the distance margin. The loss is independent of the parameterization of $\Phi, \tilde{\Phi}$ but depends on the selection of the triplets $\T=\{t_i\}$ for $\X_n$, given the orbits, and the canonical instance $x_c$ for each orbit. 

\subsection{Orbit triplet loss}

For $\lambda_2=0$ the orbit metric loss reduces to the triplet loss in Eq.~(\ref{eq:triplet_loss}), when similarity and dissimilarity are specified by orbit memberships. Points that lie on the same orbit are pulled together and points on different orbits are pushed apart. The minimizer will be pushed to satisfy Eq.~(\ref{eq:triplet_motiv}), using all triplets in the training set. Note that in the theoretical minimum of $L_t$, e.g. using the subgradient of the hinge loss, Eq.~(\ref{eq:triplet_motiv}) is satisfied. The orbit triplet loss follows a \emph{Siamese network} architecture \cite{Chopra2005}, with a tied-weight embedding trained using triplets as input. 

The following proposition shows how, for the case of bounded-norm embeddings, minimizing the triplet loss, thus pushing to minimize Eq.~(\ref{eq:triplet_motiv}), leads to an operational definition of selective robustness to transformations (invariance) with a tolerance margin $\alpha$. {\prop Let $\Phi$ be in the space of functions with norm bounded by $\sqrt{2\alpha}$, i.e. $ \Phi \in \H \subseteq \{ \Phi\,|\,\Phi\!:\!\X \rightarrow \F, \nor{\Phi}^2_{\F}\leq 2\alpha\}$. If Eq.~(\ref{eq:triplet_motiv}) is true, then $\Phi$ is invariant for the orbit transformations and selective for the orbit identities, according to the $\epsilon$-approximate definition in Eq.~(\ref{eq:inv_sel}) with $\epsilon=\alpha$.} 
\begin{proof}
For a triplet $(x, x_p, x_q)$, with $x_p, x_q$ being positive and negative examples for $x$, Eq.~(\ref{eq:triplet_motiv}) gives  $D(x, x_q) \geq D(x, x_p) + \alpha \geq \alpha, \forall x, x_p \in \X$ and $D(x, x_p) \leq D(x, x_q) - \alpha \leq \alpha$, as by the bounded norm assumption $D(x, x_q) = ||\Phi(x) - \Phi(x_q)||^2_{\R^k}\leq 2\alpha, \forall x, x_q \in \X$. Since $x_p, x_q$ are same and different orbit elements, it holds that $x \sim x_p, x \nsim x_q \iff D(x, x_p) \leq \alpha$, i.e. $\Phi$ satisfies (\ref{eq:inv_sel}) with $\epsilon=\alpha$.  
\end{proof}

\subsection{Orbit encoder}
\label{subsec:oe}

For $\lambda_1=0$ the orbit metric loss reduces to a loss that penalizes, using an additional decoder map $\tilde{\Phi}:\F \rightarrow \X$, the reconstruction error between the output of point $x_i$ to the canonical $x_c$ of the orbit $\O_{x_i}$. This is also the error of the transformation rectification that $\tilde{\Phi}\circ\Phi:\X \rightarrow \X$ applies on the input $x_i$, assumed to be the transformation of $x_c$. This loss is a novel type of autoencoder loss that learns to de-transform the input. 

The motivation is the generalization of denoising autoencoders, that learn to reconstruct clean versions of their noisy input, to transformations with or without an explicit generative model, using the equivalence of points within an orbit. \emph{Orbit encoders} learn to de-transform an input adaptively, for all transformations in the training set orbits, by mapping points onto a pre-selected canonical orbit element (Fig.~\ref{fig:orbenc}, top row). This provides a reference point for the set, such that every point in $\O_{x_i}$ can be seen as $x_i = c_i(x_c)$ or $x_i = g_ix_c$ for a known transformation process. The reconstruction error is then $||x_c - \tilde{\Phi}\circ\Phi(g_ix_c)||$ and the minimization pushes the solution towards an `inversion' of the transformation $\tilde{\Phi}\circ\Phi \approx g_i^{-1}$, jointly for all points in the training set. Another way to see the rectification objective is as trying to reconstruct any given $x_c$ from an artificially transformed version of it $x_i = g_ix_c$.  

Training requires pairs $(x_c, x_i)$ and the choice of the canonical for each orbit has to be consistent only across the same semantic class of a downstream task, e.g. all orbits of the same class. The loss enforces selectivity on $\Phi$ by preserving sufficient information to reconstruct the input irrespective of the transformation.

\subsection{Parameterization of the embedding}

The mapping $\Phi$ is parameterized through multiple layers  
\begin{equation}
\Phi = \Phi_L \circ \cdots \circ \Phi_1, \Phi_l: \F^{l-1} \rightarrow \F^{l+1}, \F^0 = \R^d, \F^L = \R^k
\end{equation}
with each one being a $k_l$-dimensional feature map of linear projections on $f_l$ filters and nonlinearities of the form in Eq.~(\ref{eq:enc_basic}). The output of layer $l$ given layer $l-1$ is 
\begin{equation}
    \Phi_l \circ \Phi_{l-1}(\cdot)  = \sigma \circ (W_l \Phi_{l-1}(\cdot) + b_l),
    \label{eq:phi_param1}
\end{equation}
where $W_l \in \R^{k_l \times k_{l-1}}$ the $k_l \times k_{l-1}$ weight matrix and $b_l \in \R^{k_l}$.
We consider convolutional maps, where groups of filters correspond to the same local kernel shifted over the support of the input, i.e. each filter is sparse on the input (local connectivity) and the projection is a convolution operator (weight sharing). The activation in Eq.~(\ref{eq:phi_param1}) is then  
\begin{equation}
    \Phi_l \circ \Phi_{l-1}(\cdot)  = \sigma \circ (W_l \star \Phi_{l-1}(\cdot) + b_l)
    \label{eq:phi_param2}
\end{equation}
where $\star$ denotes convolution with each row of $W_l \in \R^{f_l \times k_{l-1}}$ (with $k_l = f_l \times c_l$, where $c_l$ the number of shifts of the convolution kernel) and $b_l \in \R^{f_l}$ (one bias per channel). 

For the nonlinearity $\sigma$ we use the hard rectifier (ReLU activation functions) given by $\sigma(a) = |a|_{+} = \max\{0, a\}, a \in \R$, which is applied element wise on the pre-activation output $h_l = (W_l \star \Phi_{l-1}(\cdot) + b_l) \in \R^{k_l}$, i.e. $(\sigma \circ h_l)^j = \sigma(h_l^j), j=1\ldots,k_l$. Batch normalization is applied on $h_l$ before $\sigma$ as $(\eta \circ h_l)^j = \eta(h_l^j) = \gamma^j h_l^j + \beta^j$, where $\gamma$ and $\beta$ are trainable parameter vectors and each dimension of $h_l$ is standardized to be zero mean and unit variance, using the statistics of the training mini-batch \cite{Ioffe2015}. 

In addition, max pooling nonlinearities are introduced after a number of convolution layers in order to increase spatial invariance and decrease the feature map sizes. For each filter $j$, the layer is looking at the corresponding support in its input $\Phi_{l-1}\ind_{f_l=j}$ and takes the maximum over sets of convolution values defined on a grid $\mathcal{N}$ of neighboring values, i.e. $\max\{\Phi_{l-1}\ind_{\mathcal{N}, f_l=j}\}$. 

The decoder $\tilde{\Phi}$ is a series of deconvolution and un-pooling layers \cite{Zeiler2010}, in direct correspondence to the encoder in number of layers, units per layer, filters, size of kernels, and with tied weights such that $\widetilde{W_l} = W_l^T, l=1, \ldots, L$.

\begin{figure*} [th!]
\ifpgf
    \begin{center}
    \captionsetup[subfigure]{labelformat=empty}
        \begin{subfigure}{0.15\linewidth}
            \centering
           \input{figs/mnist-triplet-tsne.pgf}
        \end{subfigure}
        \begin{subfigure}{0.15\linewidth}
            \centering
           \input{figs/mnist-kitchen-sink-tsne.pgf}
        \end{subfigure}
        \begin{subfigure}{0.15\linewidth}
            \centering
           \input{figs/mnist-orbit-loss-tsne.pgf}
        \end{subfigure}
        \begin{subfigure}{0.15\linewidth}
            \centering
           \input{figs/mnist-orbit-encode-tsne.pgf}
        \end{subfigure}
        \begin{subfigure}{0.15\linewidth}
            \centering
           \input{figs/mnist-exemplar-tsne.pgf}
        \end{subfigure}
        \begin{subfigure}{0.15\linewidth}
            \centering
           \input{figs/mnist-autoencode-tsne.pgf}
        \end{subfigure}
    \end{center}
\else
    \centerline{\hfill
        \includegraphics[width=0.15\linewidth,page=4]{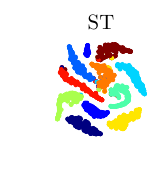}\hfill
        \includegraphics[width=0.15\linewidth,page=5]{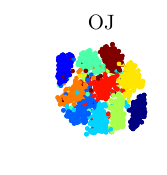}\hfill
        \includegraphics[width=0.15\linewidth,page=6]{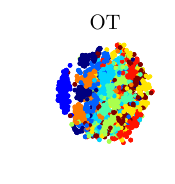}\hfill
        \includegraphics[width=0.15\linewidth,page=7]{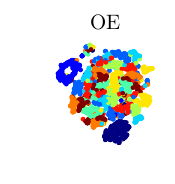}\hfill
        \includegraphics[width=0.15\linewidth,page=8]{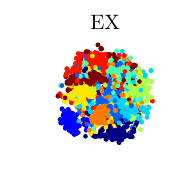}\hfill
        \includegraphics[width=0.15\linewidth,page=9]{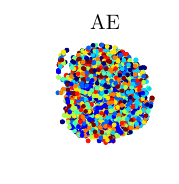}\hfill}
\fi    
\capmar  \capmar
\caption{\small 2D t-SNE visualizations of a subset of the affine MNIST test set (original set augmented by random affine transformations), under different embeddings. Colors code the 10 different semantic classes (digit identity).}
\label{fig:tsne-mnist}
\end{figure*}

\section{Experiments}

We compared the embeddings learned using the proposed loss, Orbit Joint (OJ) in Eq.~(\ref{eq:orbit_loss}) and the two special cases, Orbit Triplet (OT) ($\lambda_2=0$) and Orbit Encode (OE) ($\lambda_1=0$), and three reference, closely related methods: Supervised Triplet (ST) \cite{Schroff2015}, Exemplar (EX) \cite{Dosovitskiy2015}, and standard Autoencoder (AE). Each loss was used for learning a map from the input to a metric space using the same network/parametrization and varying degrees of supervision (unsupervised, supervised or weakly-supervised using the set of orbits). Once the embeddings were learned, the \emph{training and test sets} for the downstream tasks (one-shot digit classification on affine MNIST, face verification and retrieval on Multi-PIE) were encoded and used to evaluate the performance. The \emph{embedding set} used for training the networks, i.e. the collection of orbits, was kept separate from any data used in the downstream tasks. For the affine-MNIST evaluations, we also compared our methods to an embedding parametrization featuring a Spatial Transformer Networks module \cite{Jaderberg2015}, trained with orbit supervision (OT-STN) or full supervision (ST-STN). 

\subsection{Network and training details}

The encoder was a deep convolutional network following the VGG architecture \cite{Simonyan2015}. Each layer was a series of convolutions with a small $3 \times 3$ kernel (of stride 1, padding 1), batch normalization \cite{Ioffe2015} and ReLU activation. A spatial max pooling layer (of stride 2 and size either $2 \times 2$ or $4 \times 4$) was used every two such layers of convolutions. The number of channels doubled after each max pooling layer, ranging from 16 to 128 for MNIST and 64 to 512 for Multi-PIE. Four iterations of convolution and pooling were followed by a final fully-connected layer of size 1024. The decoder was a deconvolutional network, reversing the series of operations in the encoder using convolutional reconstruction and max unpooling \cite{Zeiler2010}. Encoder and decoder weights were tied with free biases. Training was done with minibatch Stochastic Gradient Descent using the ADAM optimizer. For MNIST experiments we used minibatches of 256, and for Multi-PIE, 72. The selection of the triplets followed the soft negative selection process from \cite{Schroff2015}. The values for $\lambda_1$ and $\lambda_2$ were set equal in these experiments, but they can be selected by cross-validation. The STNs modules, used for the affine-MNIST comparisons, consisted of two max pooling-convolution-ReLU blocks with 20 filters of size $5 \times 5$ (stride 1), pooling regions of size $2 \times 2$ and no overlap, and followed by two linear layers.

\begin{figure}
    \begin{center}
    \input{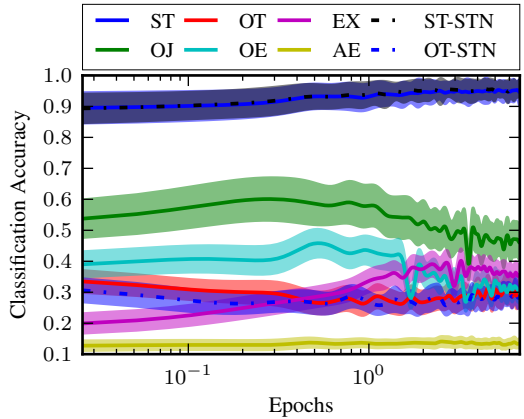}
    \end{center}
    \capmar \capmar
    \caption{\small Nearest Neighbor classification using one example-per-class on MNIST augmented with affine transformations. Accuracies are averaged over 100 random re-samples of the training set (shaded region 1 s.d.) and shown at different times during the training of the embeddings.}    
\label{fig:acc}
\end{figure}

\newcommand{\fl}{.1\linewidth}


\newcommand{\trimmedgraphic}[2][]{%
    \includegraphics[trim = 0cm 0cm 0cm 0cm, clip, width=1\linewidth,#1]%
        {#2}%
}

\begin{figure}[!t]
\begin{center}

\captionsetup[subfigure]{labelformat=empty}
\begin{subfigure}{\fl}
  \centering
 \trimmedgraphic{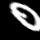}
\end{subfigure}
\begin{subfigure}{\fl}
  \centering
 \trimmedgraphic{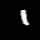}
\end{subfigure}%
\begin{subfigure}{\fl}
  \centering
 \trimmedgraphic{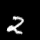}
\end{subfigure}%
\begin{subfigure}{\fl}
  \centering
 \trimmedgraphic{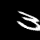}
\end{subfigure}%
\begin{subfigure}{\fl}
  \centering
 \trimmedgraphic{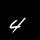}
\end{subfigure}%
\begin{subfigure}{\fl}
  \centering
 \trimmedgraphic{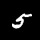}
\end{subfigure}%
\begin{subfigure}{\fl}
  \centering
 \trimmedgraphic{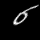}
\end{subfigure}%
\begin{subfigure}{\fl}
  \centering
 \trimmedgraphic{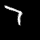}
\end{subfigure}%
\begin{subfigure}{\fl}
  \centering
 \trimmedgraphic{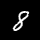}
\end{subfigure}%
\begin{subfigure}{\fl}
  \centering
 \trimmedgraphic{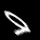}
\end{subfigure}%
\vspace{-0.1cm}
\begin{subfigure}{\fl}
  \centering
 \trimmedgraphic{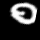}
\end{subfigure}%
\begin{subfigure}{\fl}
  \centering
 \trimmedgraphic{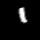}
\end{subfigure}%
\begin{subfigure}{\fl}
  \centering
 \trimmedgraphic{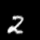}
\end{subfigure}%
\begin{subfigure}{\fl}
  \centering
 \trimmedgraphic{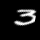}
\end{subfigure}%
\begin{subfigure}{\fl}
  \centering
 \trimmedgraphic{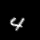}
\end{subfigure}%
\begin{subfigure}{\fl}
  \centering
 \trimmedgraphic{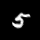}
\end{subfigure}%
\begin{subfigure}{\fl}
  \centering
 \trimmedgraphic{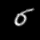}
\end{subfigure}%
\begin{subfigure}{\fl}
  \centering
 \trimmedgraphic{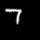}
\end{subfigure}%
\begin{subfigure}{\fl}
  \centering
 \trimmedgraphic{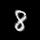}
\end{subfigure}%
\begin{subfigure}{\fl}
  \centering
 \trimmedgraphic{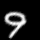}
\end{subfigure}%
\vspace{-0.1cm}
\begin{subfigure}{\fl}
  \centering
 \trimmedgraphic{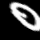}
\end{subfigure}%
\begin{subfigure}{\fl}
  \centering
 \trimmedgraphic{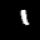}
\end{subfigure}%
\begin{subfigure}{\fl}
  \centering
 \trimmedgraphic{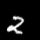}
\end{subfigure}%
\begin{subfigure}{\fl}
  \centering
 \trimmedgraphic{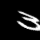}
\end{subfigure}%
\begin{subfigure}{\fl}
  \centering
 \trimmedgraphic{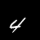}
\end{subfigure}%
\begin{subfigure}{\fl}
  \centering
 \trimmedgraphic{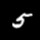}
\end{subfigure}%
\begin{subfigure}{\fl}
  \centering
 \trimmedgraphic{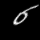}
\end{subfigure}%
\begin{subfigure}{\fl}
  \centering
 \trimmedgraphic{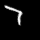}
\end{subfigure}%
\begin{subfigure}{\fl}
  \centering
 \trimmedgraphic{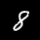}
\end{subfigure}%
\begin{subfigure}{\fl}
  \centering
 \trimmedgraphic{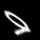}
\end{subfigure}%
\vspace{-0.1cm}
\begin{subfigure}{\fl}
  \centering
 \trimmedgraphic{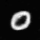}
\end{subfigure}%
\begin{subfigure}{\fl}
  \centering
 \trimmedgraphic{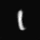}
\end{subfigure}%
\begin{subfigure}{\fl}
  \centering
 \trimmedgraphic{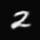}
\end{subfigure}%
\begin{subfigure}{\fl}
  \centering
 \trimmedgraphic{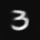}
\end{subfigure}%
\begin{subfigure}{\fl}
  \centering
 \trimmedgraphic{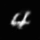}
\end{subfigure}%
\begin{subfigure}{\fl}
  \centering
 \trimmedgraphic{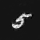}
\end{subfigure}%
\begin{subfigure}{\fl}
  \centering
 \trimmedgraphic{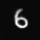}
\end{subfigure}%
\begin{subfigure}{\fl}
  \centering
 \trimmedgraphic{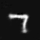}
\end{subfigure}%
\begin{subfigure}{\fl}
  \centering
 \trimmedgraphic{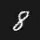}
\end{subfigure}%
\begin{subfigure}{\fl}
  \centering
 \trimmedgraphic{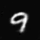}
\end{subfigure}%
\end{center}
\capmar \capmar
\caption{\small 
Input (row 1) and rectifications for ST-STN (row 2), OT-STN (row 3) and OJ (row 4). STNs produced convincing rectifications when full supervision was available (ST-STN) but learned the identity with weak supervision (OT-STN). The encoder-decoder map trained with the proposed orbit metric loss produces accurate rectifications using only weak supervision.}
\label{fig:stn-rec}
\end{figure}

\begin{figure*}[!t]
\ifpgf
    \begin{center}
    \captionsetup[subfigure]{labelformat=empty}
    \begin{subfigure}{0.15\linewidth}
      \centering
      \input{figs/mpie-triplet-tsne.pgf}
    \end{subfigure}
    \begin{subfigure}{0.15\linewidth}
      \centering
      \input{figs/mpie-kitchen-sink-tsne.pgf}
    \end{subfigure}
    \begin{subfigure}{0.15\linewidth}
      \centering
       \input{figs/mpie-orbit-loss-tsne.pgf}
    \end{subfigure}
    \begin{subfigure}{0.15\linewidth}
      \centering
      \input{figs/mpie-orbit-encode-tsne.pgf}
    \end{subfigure}
    \begin{subfigure}{0.15\linewidth}
      \centering
      \input{figs/mpie-exemplar-tsne.pgf}
    \end{subfigure}
    \begin{subfigure}{0.15\linewidth}
      \centering
        \includegraphics[]{figs_png/mpie-autoencode-tsne.png}
    \end{subfigure}
    \end{center}
\else
    \centerline{\hfill
        \includegraphics[width=0.15\textwidth,page=10]{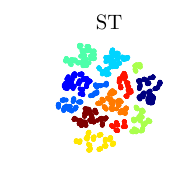}\hfill
        \includegraphics[width=0.15\textwidth,page=11]{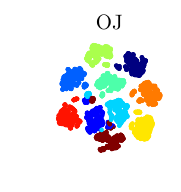}\hfill
        \includegraphics[width=0.15\textwidth,page=12]{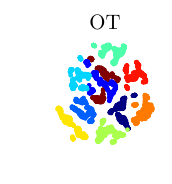}\hfill
        \includegraphics[width=0.15\textwidth,page=13]{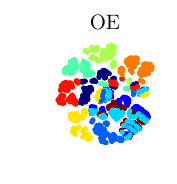}\hfill
        \includegraphics[width=0.15\textwidth,page=14]{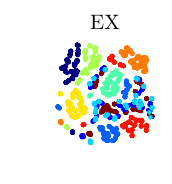}\hfill
        \includegraphics[width=0.15\textwidth,page=15]{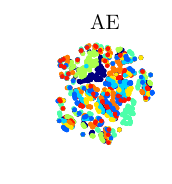}\hfill}
\fi
\capmar \capmar
\caption{\small 2D t-SNE visualizations of the embeddings for face images from 10 subjects/classes (coded by different colors), randomly sampled from Multi-PIE test set.}
\label{fig:tsne-mpie}
\end{figure*}

\begin{figure*}[!ht]
    \begin{center}
        \captionsetup[subfigure]{labelformat=empty}
        \begin{subfigure}{\fl}
            \centering
            \trimmedgraphic{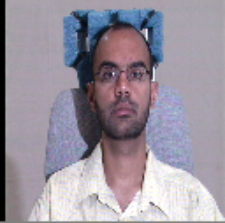}
        \end{subfigure}%
        \begin{subfigure}{\fl}
            \centering
            \trimmedgraphic{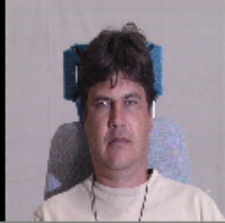}
        \end{subfigure}%
        \begin{subfigure}{\fl}
            \centering
            \trimmedgraphic{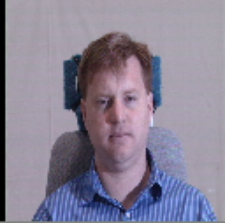}
        \end{subfigure}%
        \begin{subfigure}{\fl}
            \centering
            \trimmedgraphic{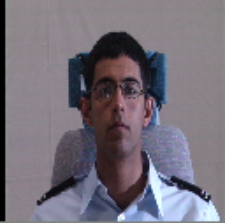}
        \end{subfigure}%
        \begin{subfigure}{\fl}
            \centering
            \trimmedgraphic{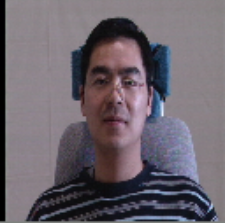}
        \end{subfigure}%
        \begin{subfigure}{\fl}
            \centering
            \trimmedgraphic{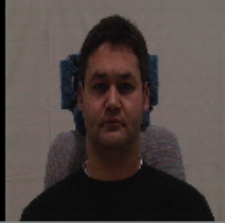}
        \end{subfigure}%
        \begin{subfigure}{\fl}
            \centering
            \trimmedgraphic{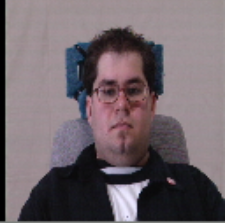}
        \end{subfigure}%
        \begin{subfigure}{\fl}
            \centering
            \trimmedgraphic{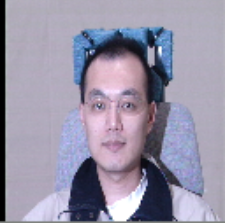}
        \end{subfigure}%

        \begin{subfigure}{\fl}
            \centering
            \trimmedgraphic{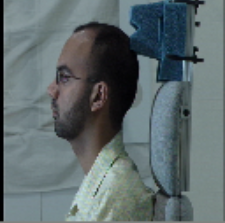}
        \end{subfigure}%
        \begin{subfigure}{\fl}
            \centering
            \trimmedgraphic{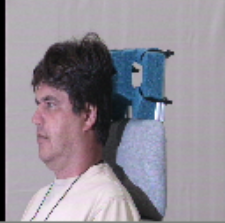}
        \end{subfigure}%
        \begin{subfigure}{\fl}
            \centering
            \trimmedgraphic{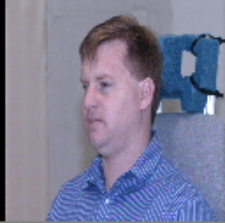}
        \end{subfigure}%
        \begin{subfigure}{\fl}
            \centering
            \trimmedgraphic{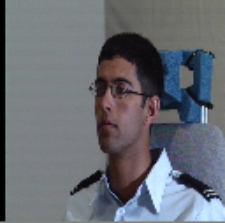}
        \end{subfigure}%
        \begin{subfigure}{\fl}
            \centering
            \trimmedgraphic{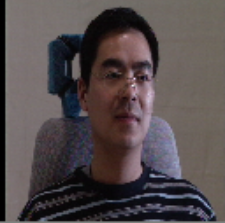}
        \end{subfigure}%
        \begin{subfigure}{\fl}
            \centering
            \trimmedgraphic{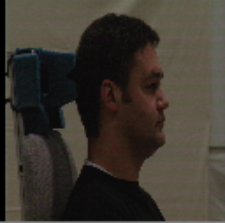}
        \end{subfigure}%
        \begin{subfigure}{\fl}
            \centering
            \trimmedgraphic{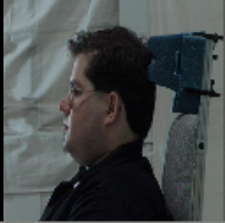}
        \end{subfigure}%
        \begin{subfigure}{\fl}
            \centering
            \trimmedgraphic{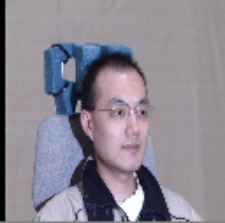}
        \end{subfigure}%

        \begin{subfigure}{\fl}
            \centering
            \trimmedgraphic{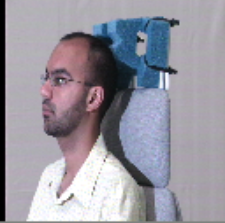}
        \end{subfigure}%
        \begin{subfigure}{\fl}
            \centering
            \trimmedgraphic{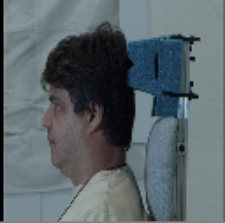}
        \end{subfigure}%
        \begin{subfigure}{\fl}
            \centering
            \trimmedgraphic{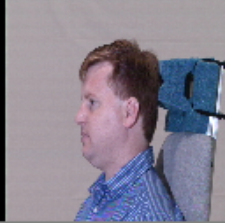}
        \end{subfigure}%
        \begin{subfigure}{\fl}
            \centering
            \trimmedgraphic{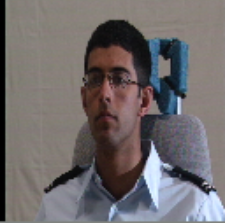}
        \end{subfigure}%
        \begin{subfigure}{\fl}
            \centering
            \trimmedgraphic{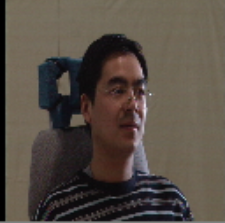}
        \end{subfigure}%
        \begin{subfigure}{\fl}
            \centering
            \trimmedgraphic{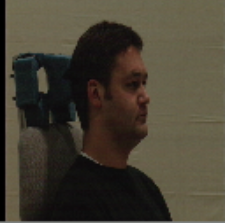}
        \end{subfigure}%
        \begin{subfigure}{\fl}
            \centering
            \trimmedgraphic{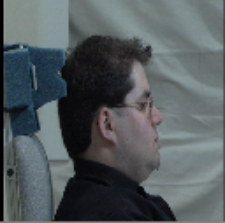}
        \end{subfigure}%
        \begin{subfigure}{\fl}
            \centering
            \trimmedgraphic{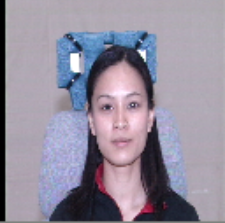}
        \end{subfigure}%

        \begin{subfigure}{\fl}
            \centering
            \trimmedgraphic{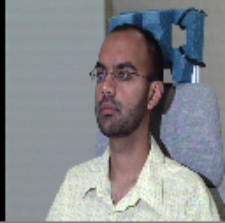}
        \end{subfigure}%
        \begin{subfigure}{\fl}
            \centering
            \trimmedgraphic{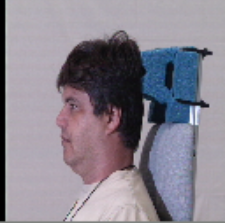}
        \end{subfigure}%
        \begin{subfigure}{\fl}
            \centering
            \trimmedgraphic{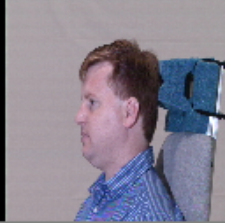}
        \end{subfigure}%
        \begin{subfigure}{\fl}
            \centering
            \trimmedgraphic{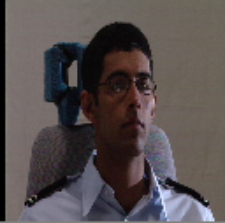}
        \end{subfigure}%
        \begin{subfigure}{\fl}
            \centering
            \trimmedgraphic{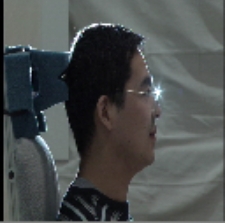}
        \end{subfigure}%
        \begin{subfigure}{\fl}
            \centering
            \trimmedgraphic{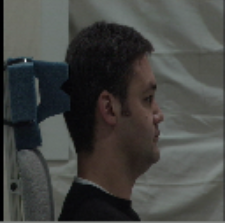}
        \end{subfigure}%
        \begin{subfigure}{\fl}
            \centering
            \trimmedgraphic{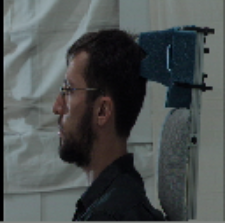}
        \end{subfigure}%
        \begin{subfigure}{\fl}
            \centering
            \trimmedgraphic{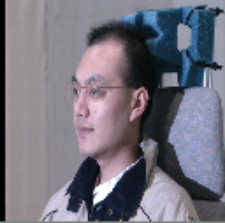}
        \end{subfigure}%

        \begin{subfigure}{\fl}
            \centering
            \trimmedgraphic{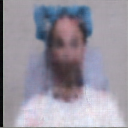}
        \end{subfigure}%
        \begin{subfigure}{\fl}
            \centering
            \trimmedgraphic{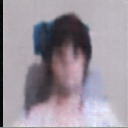}
        \end{subfigure}%
        \begin{subfigure}{\fl}
            \centering
            \trimmedgraphic{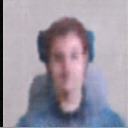}
        \end{subfigure}%
        \begin{subfigure}{\fl}
            \centering
            \trimmedgraphic{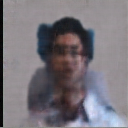}
        \end{subfigure}%
        \begin{subfigure}{\fl}
            \centering
            \trimmedgraphic{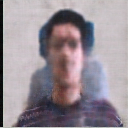}
        \end{subfigure}%
        \begin{subfigure}{\fl}
            \centering
            \trimmedgraphic{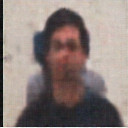}
        \end{subfigure}%
        \begin{subfigure}{\fl}
            \centering
            \trimmedgraphic{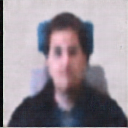}
        \end{subfigure}%
        \begin{subfigure}{\fl}
            \centering
            \trimmedgraphic{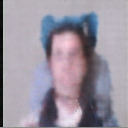}
        \end{subfigure}%
    \end{center}
    \caption{\small Examples of retrieval and rectification on Multi-PIE. Rows correspond to: 1) canonical pose of the seec image, 2) seed image, 3) Top-1 retrieved using the Orbit Joint (OJ) embedding, 4) Top-1 retrieved using the Exemplar (EX) embedding \cite{Dosovitskiy2015} and 5) the reconstructed canonical pose using the decoder of the OJ network.}
    \label{fig:mpie-data}
\end{figure*}

\subsection{Affine transformations: MNIST}
We created a version of MNIST, using 32 random affine transformation for each point in the original MNIST dataset (samples in Fig.~\ref{fig:orbenc}, middle row). Transformations were sampled uniformly from the union of the following intervals: rotation from $[-90^{\circ},90^{\circ}]$, shearing factor from $[-0.3, 0.3]$, scale factor from $[0.7, 1.3]$, and translation in each dimension from $[-15,15]$ pixels. The orbit set consisted of the original MNIST training set ($50 \times 10^3$ images), augmented by 32 transformations for each sample, resulting in a total of $1650 \times 10^3$ images, grouped in $50 \times 10^3$ orbits. Each orbit, of size 33, is the set of a single original image (canonical) and the corresponding random transformations of it. 

The learned embeddings using this set were employed in a one-shot classification task to assess their invariance and selectivity properties. The training set consisted of 10 images, one from each semantic class. These were drawn at random from the original MNIST validation (augmented by 32 random affine transformations). The test set consisted of $25 \times 10^3$ images randomly drawn from the original MNIST test set (plus transformations). Figure~\ref{fig:tsne-mnist} shows the 2D t-SNE plots \cite{Maaten2008} of the learned embeddings on a random subset of the test set. Qualitatively, the best separation and grouping was observed with the fully supervised triplet loss (ST), followed by the weakly supervised orbit joint loss (OJ).  

Nearest Neighbor classification was used for predicting the label of each test point from the 10 image labelled training set, which is not controlled for transformations. Figure~\ref{fig:acc} shows classification accuracy results during embedding training epochs. At each iteration, the accuracy is shown as mean with standard deviation (sdtv) error bars over 100 different labelled set selections. As expected, the supervised ST performed best and the unsupervised AE was the lower baseline. Of the weakly-supervised methods, the orbit metric loss OJ achieved the top accuracy, followed by OE and EX. Spatial transformer network modules provided a small improvement in accuracy ($2\%$, consistent with the improvement reported in \cite{Jaderberg2015}) when used with full supervision (ST vs. ST-STN). However, when only orbit information was available (OT vs. OT-STN), there was no difference in performance. This is further reflected in Fig.~\ref{fig:stn-rec} which shows rectification examples from the output of the STN module and the learned decoder with our method.

\subsection{Face transformations: Multi-PIE}

The Multi-PIE dataset \cite{Gross2010} contains images of faces of 129 individuals, captured from 13 distinct viewpoints and under 20 different illumination conditions. Acquisition was carried out across four sessions, resulting in a dataset of $129\times13\times20\times4 = 134160$ images. For learning the maps from the input to the metric space, we used all images from three of the sessions to form the embedding sets and left out all images from the fourth session for performance assessment in the downstream task. During training of the embedding map $\Phi$, the ST method had access to the face identity for each image, thus considering all images of the same subject (across sessions, viewpoints and illumination conditions) as belonging to the same equivalence class set. The weakly supervised methods (OJ, OT, OE, EX) on the other hand, had only access to the set of orbits formed by partitioning the embedding set in $129 \times 20 \times 3$ orbits, each corresponding to all 13 viewpoints for a single identity, illumination condition and session (Fig.~\ref{fig:orbex}, row 2).

For the purpose of performance assessment, we used the learned maps $\Phi$ and encoded the held-out test set (one session). Figure~\ref{fig:tsne-mpie} shows the relative distance landscape of the learned embeddings, as 2D t-SNE plots, for all images from 10 subjects of the test set. The weakly supervised OJ appears to have similar or better grouping and separability properties than the fully supervised ST. We used two distance-based tasks to quantitatively evaluate the embedding metric spaces: a \textit{same-different} face verification task and a face retrieval task. In a transformation-robust metric space for face representation, same-identity images should be closer to each other than to other identities, and the nearest neighbor to each should be an image of the same class. We measure the Area Under the ROC Curve (AUC) for verification 
and the mean top-1 precision for retrieval.
The process of training and evaluating an embedding was repeated on all four possible 3-1 splits, across sessions of Multi-PIE, to assess the uncertainty in the performance measures.

For the verification task, we used all unique pairwise distances in the embedding space, considered all possible decision thresholds and integrated the True Positive and False Positive rates to compute the AUC. For the retrieval task, we select the closest point to a query image (top-1 retrieval) from a target search set. We considered each test image individually as query, using the rest of the test set as the target set, after removing all same-identity images (32 in total, including the query) at the same illumination (regardless of viewpoint) and at the same viewpoint (regardless of illumination). This made for a more challenging task and helped in ensuring that the embeddings are evaluated with respect to their preference of identity over appearance, e.g. by excluding candidates with strong pose or illumination bias. As a performance measure, we report the mean precision, i.e. the fraction of queries that yielded a correct retrieval. 

\begin{figure}[!t!]
    \begin{center}
        \input{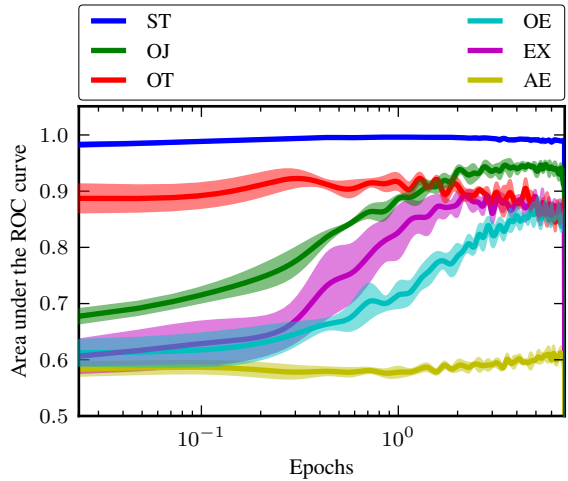}
    \end{center}
    \capmar \capmar
    \caption{\small Verification performance on Multi-PIE with epochs of embedding training: Mean Area Under the ROC Curve, with 1 s.d. error bars, over 4 different splits (3 sessions for embedding training, 1 for evaluation).}
\label{fig:auc}
\end{figure}

\begin{figure}[!t!]
    \begin{center}
        \input{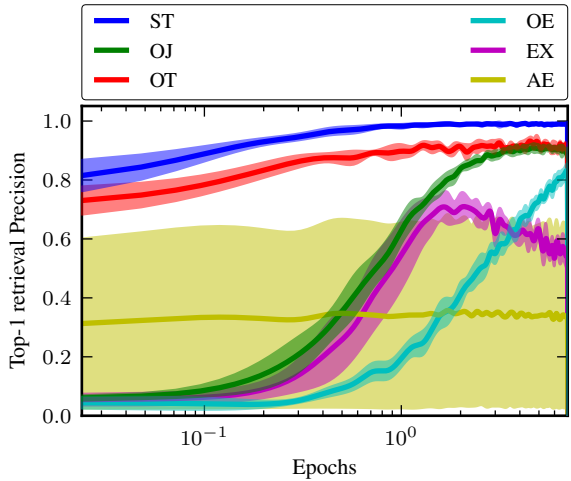}
    \end{center}
    \capmar \capmar
    \caption{\small Retrieval performance on Multi-PIE with epochs of embedding training: Mean top-1 precision, with 1 s.d. error bars, over 4 different splits (3 sessions for embedding training, 1 for evaluation).}
\label{fig:retrieval}
\end{figure}

\begin{table*}
    \begin{center}
        \def\arraystretch{0.9}
        \begin{tabular}{@{}r@{ }||c |c c|c c|c c r@{}}
        \toprule
            {} & {\small OJ} &{\small EX} \cite{Dosovitskiy2015}  & {\small $p$-val} & {\small OT} & {\small $p$-val} & {\small OE} & {\small $p$-val} \\                        
            \midrule 
            {\small Multi-PIE \textbf{AUC}} & {\small $0.94\!\pm\!0.012$}& {\small $0.90\!\pm\! 0.007$} & {\small $\mathbf{6e\!-\!03}$} & {\small $0.93\!\pm\!0.008$} & {\small $6e\!-\!01$} & {\small$0.87\!\pm\!0026$} & {\small $\mathbf{4e\!-\!02}$} & \\
            {\small  \textbf{Top-1}} &  {\small $0.95\!\pm\!0.029$} &{\small $0.80\!\pm\!0.035   $} & {\small $\mathbf{2e\!-\!02}$} & {\small $0.97\!\pm\!0.004$} & {\small $8e\!-\!01$} & {\small $0.99\!\pm\!0.002$} & {\small $8e\!-\!02$} &\\
            \midrule 
            {\small Affine MNIST \textbf{ACC}} & {\small $0.66\!\pm\! 0.027$} & {\small $0.45\!\pm\! 0.019$} & {\small $\mathbf{2e\!-\!09}$} & {\small $0.35\!\pm\!0.021$} & {\small $\mathbf{2e\!-\!09}$} & {\small $0.49\!\pm\!0.011$}  & {\small $\mathbf{6e\!-\!08}$} &\\
            \bottomrule
        \end{tabular}        
    \end{center}
    \capmar
    \caption{\small Generalization using embedding-validation-test splits and independent selection of the training stopping time using the validation set. $P$-values quantify significance of the difference between OJ-EX, OJ-OT, and OJ-OE (bolded if $p < 0.05$). 
    }\label{tab:pvalue}
\end{table*}

Verification performance is shown in Fig.~\ref{fig:auc} as mean and s.d. of AUC across 3-1 splits of Multi-PIE sessions (3 for embedding training--1 for evaluation). As expected, the weakly supervised methods, that access only the orbit assignments for training, are in-between the ST loss, which has access to category-level labels and the unsupervised AE loss. Orbit triplet (OT) learns very quickly and its performance tends to decrease after a few iterations. The other methods learn at comparable rates. The joint orbit loss (OJ) achieves the best AUC score. A similar ranking holds for the retrieval task, shown as top-1 precision in Fig.~\ref{fig:retrieval}.

\subsection{Early stopping by cross validation}

To evaluate the generalization performance, i.e. testing on a set of unseen examples, we trained all embeddings by applying cross-validation for individually selecting the number of iterations. We compared the proposed OJ to the state-of-the-art, weakly-supervised EX loss using six splits (4 choose 2) by session on Multi-PIE (2-Embedding -- 1-Validation (VA) -- 1-Test (TE)) and 10 random splits of the MNIST test set (VA -- TE), with elements of the same orbit appearing in only one. We selected the stopping time that gave the best performance measure (AUC, top-1 precision, accuracy) on the VA set and evaluated the same measure on the TE set. Table~\ref{tab:pvalue} shows the mean and s.d. over splits, with corresponding p-values (paired t-test with Bonferroni correction) quantifying significance for the difference between OJ-EX, OJ-OT and OJ-OE. OJ consistently outperforms EX in all three generalization tasks. Moreover, the performance of OJ is either better or statistically indistinguishable (with a standard significance threshold at $p < 0.05$) from OT and OE. This observation makes the case for the joint loss, which can result in substantial improvements like in the one-shot classification task on affine MNIST. Furthermore, it suggests that careful selection of the relative weights ($\lambda_1$ and $\lambda_2$) of the triplet and reconstruction terms in the OJ loss (Eq.~\ref{eq:orbit_loss}), e.g. via cross-validation, could be beneficial.

\section{Conclusions}

We introduced a loss function that combines a \emph{discriminative} and a \emph{generative} term for learning embeddings, using weak supervision from generic transformation orbits. We showed that the resulting image embeddings induce a metric space that is relevant for distance-based learning tasks such as one-shot learning classification, face verification and retrieval.
The two loss terms serve complementary purposes, so that joint training is advantageous and supersedes state-of-the-art, exemplar-based training and, when applicable Spatial Transformer Networks. Transformations that do not alter the semantic category of the input are present in most classical perception problems, from pitch shifts in speech recognition to pose, illumination and gait changes in action recognition to reflectance properties for object categorization. The work presented here suggests that explicitly defining equivalence classes according to these transformations is a rich, weak supervision signal that can be exploited in a more general class of representation learning methods, starting from the proposed loss function, to learn semantically relevant embeddings. Such embeddings define distance functions that are robust to typical transformations and are useful for categorization, retrieval, verification and clustering. Future work should assess the relevance for other modalities, such as video or audio and the potential of acquiring the equivalence classes through time continuity.

\section*{Acknowledgements}
{\small
\noindent This material is based upon work supported by the Center for Brains, Minds and Machines (CBMM), funded by NSF STC award CCF-1231216. The DGX-1 used for this research was donated by the NVIDIA Corporation. Stephen Voinea acknowledges the support of a Nuance Foundation Grant. The authors gratefully acknowledge Tomaso Poggio and Fabio Anselmi for insightful discussions.}

{\small
    \bibliographystyle{ieee}
    \bibliography{OrbitMetric}
}

\newpage

\appendix
\section{Appendix: Decoder output examples}

\begin{figure}[!b!]
    \begin{center}
        \includegraphics[width=0.5\linewidth]{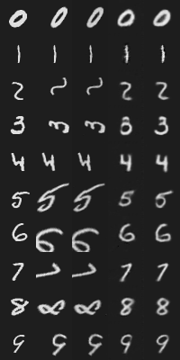}
    \end{center}
    \capmar \capmar
    \caption{\small 
    \textbf{Column 1}: canonical image, chosen randomly from the MNIST test set. \textbf{Column 2}: seed image generated from the canonical via a random affine transformation. \textbf{Columns 3-5} show the result of passing the seed image through the decoders of the Autoencoder, Orbit Encode and Orbit Joint trained networks respectively.}
    \label{fig:mnist-large}
\end{figure}

\begin{figure}
    \begin{center}
        \includegraphics[width=0.85\linewidth]{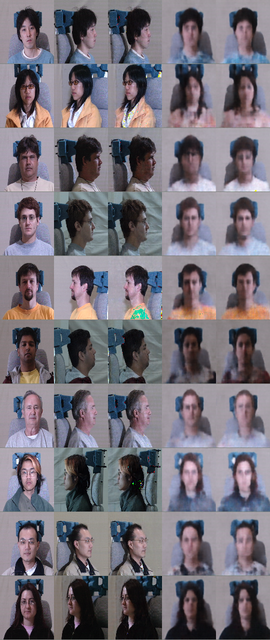}
    \end{center}
    \capmar \capmar
    \caption{\small 
    \textbf{Column 1}: canonical image, chosen randomly from the Multi-PIE test set. \textbf{Column 2}: seed image generated from the canonical via a random affine transformation. \textbf{Columns 3-5} show the result of passing the seed image through the decoders of the Autoencoder, Orbit Encode and Orbit Joint trained networks respectively.}
    \label{fig:mpie-large}
\end{figure}

Figure~\ref{fig:orbenc} and Fig.~\ref{fig:mpie-data} showed examples of rectifications applied on $g_ix_c$, where $x_c$ is a seed, untransformed image, and $g_i$ an latent transformation, using the output of the decoder $\tilde{\Phi} \circ \Phi(g_ix)$. While for OE, this rectification is the sole criterion that drives the learned embedding, for OJ, this competes with the triplet loss term. This will result in different learned features, and image outputs from the decoder. 

The effect of the joint training in the visual appearance of the rectified outputs is shown for ten images from affine MNIST dataset in Fig.~\ref{fig:mnist-large} and ten images from Multi-PIE in Fig.~\ref{fig:mpie-large}. Each one depicts the decoder output for standard AE (column 3), OE (column 4) and OJ (column 5). The autoencoder output is included as a sanity check of the reconstruction loss, i.e., the output is a faithful reconstruction of the input, which includes the transformation effect. The decoders for the OE and OJ losses both do well at rectifying the transformation (affine for MNIST and 3D viewpoint for Multi-PIE), i.e. mapping the transformed input (column 2) to an image that resembles the untransformed one (column 1). One can note subtle differences in the outputs, e.g., the 3 and 9 instances in MNIST but particularly in Multi-PIE, though it is not easy to select one based on visual qualities.

\end{document}